\documentclass[10pt, a4paper]{article}

\usepackage{lrec-coling2024_templates/lrec-coling2024} 

\usepackage{times}
\usepackage{latexsym}
\usepackage{textcomp}
\usepackage{booktabs}
\usepackage{subcaption}
\usepackage{fancyhdr}
\usepackage[bottom]{footmisc}
\usepackage[T1]{fontenc}
\usepackage[scaled]{helvet}
\usepackage[utf8]{inputenc}
\usepackage{microtype}
\usepackage{inconsolata}
\usepackage{tikz-dependency}
\usepackage{wrapfig}
\usepackage{graphicx}
\graphicspath{ {./prodigy/} }
\usepackage{makecell}
\usepackage{tabularx}
\usepackage{multirow}
\usepackage{adjustbox}
\usepackage{mathtools}
\usepackage{enumitem}
\usepackage{algorithm2e}
\usepackage{csquotes}
\usepackage{courier}
\usepackage{tcolorbox}
\usepackage{soul}
\usepackage{setspace}
\usepackage{caption}
\usepackage{multicol}
\usepackage{stfloats}
\usepackage[section]{placeins}

\usepackage{todonotes}

\title{\textbf{An Analysis of Sentential Neighbors in Implicit Discourse Relation Prediction}}

\name{Evi Judge, Reece Suchocki, Konner Syed} 

\address{The University of Colorado, Boulder \\
         Evi.Judge@Colorado.edu, Reece.Suchocki@Colorado.edu, \\ Konner.Syed@Colorado.edu\\
}

\abstract{ Discourse relation classification is an especially difficult task without explicit context markers \cite{Prasad2008ThePD}. Current approaches to implicit relation prediction solely rely on two neighboring sentences being targeted, ignoring the broader context of their surrounding environments \cite{Atwell2021WhereAW}. In this research, we propose three new methods in which to incorporate context in the task of sentence relation prediction: (1) Direct Neighbors (DNs), (2) Expanded Window Neighbors (EWNs), and (3) Part-Smart Random Neighbors (PSRNs). Our findings indicate that the inclusion of context beyond one discourse unit is harmful in the task of discourse relation classification. 
 \\ \newline  } 

\begin{document}

\maketitleabstract








\section{Introduction}

In natural language processing, discourse relation classification is the task of labeling semantic relations between text spans. Without explicitly provided context, such as connective words (e.g., \textit{since}, \textit{however}, \textit{as a result}, etc.), labeling semantic relations must rely on more subtly embedded context \citep{Prasad2008ThePD}. Discourse relation classification is important for downstream tasks such as machine translation, natural language generation, and information extraction.

\begin{figure}[h]
    \centering
    \includegraphics[width=1.0\columnwidth]{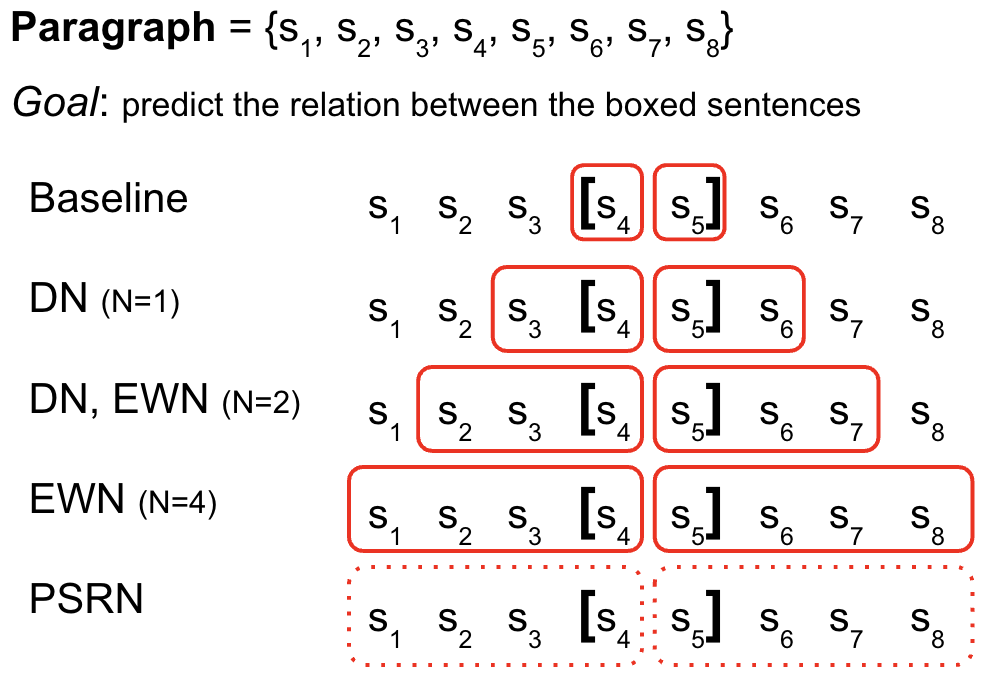}
    \caption{Expanded Context Windows.}
    \label{fig:1}
\end{figure}

A number of strategies have been proposed to improve upon discourse relation classification. \citet{Atwell2021WhereAW} suggests that context beyond the two target sentences could be beneficial to this task, but it is unclear how much context a model would need to optimally predict discourse relations. In other words, discourse pairs of any target relation may not be enough context to disambiguate and correctly classify the relation between the two discourse units. Our approach to discourse relation classification accounts for context beyond the immediate discourse unit pairs being classified. These strategies are illustrated in Figure \ref{fig:1}.


\section{Related Work}

The Penn Discourse TreeBank offers an alternative way to model the task of discourse relation prediction by focusing on local coherence relations and not document-level relations \citep{Prasad2008ThePD}. The corpus presents annotations for both explicit and implicit discourse relations. Early work was focused on the easier task of explicit relation prediction \citep{Pitler2008EasilyID}, then \citet{Pitler2009AutomaticSP} were the first to attempt predicting implicit discourse relations. \citet{Pitler2009AutomaticSP} noted that context is important in predicting implicit discourse relations.

Later, \citet{Dai2018ImprovingID} examined implicit discourse relation classification in PDTB-2.0 using compositional meaning from word interactions in word2vec embeddings and attention mechanisms in a BiLSTM-based approach \citep{Mikolov2013DistributedRO}. In their work, \citet{Dai2018ImprovingID} recognize the importance of paragraph-level context and situate their discourse units, which are typically defined as a clause or sentence, within the context of the entire paragraph citing the following reasons: (1) implicit discourse relations might require a wider understanding than simply looking at adjacent discourse units, and (2) any given discourse unit might be involved with multiple discourse relations. In the same year, \citet{Bai2018DeepER} proposed a neural model augmented by word-, sentence-, and pair- level modules to help predict implicit discourse relations.

Large language models (LLMs), such as BERT and RoBERTa, have also been applied to the task of implicit relation prediction (e.g., \citealp{devlin-etal-2019-bert}, \citealp{Liu2020OnTI}, and \citealp{Shi2019NextSP}). Results using LLMs are encouraging with BERT-based models outperforming existing systems by 5\% accuracy \citep{Shi2019NextSP}, and a module-based RoBERTa achieving accuracy above 70\% (F1>60\%) on the four-way classification task on PDTB's \textit{Class} relations \citep{Liu2020OnTI}.

In another approach, \citet{Wu2021ALD} created a model with a graph convolutional network encoder and top-down decoder in order to predict implicit discourse relations. They also introduce an auxiliary decoder that aids the main decoder in capturing complementary bidirectional label dependence.
This previous work, especially \citet{Dai2018ImprovingID}, motivates the current research and our aim to explore whether or not greater sentential context plays a role in implicit discourse relation prediction.

\section{Data}

We use the Penn Discourse TreeBank (PDTB), which is available at the \href{https://catalog.ldc.upenn.edu/LDC2008T05}{LDC}.

\subsection{The Penn Discourse TreeBank 2.0}

\begin{figure}[h]
    \hspace*{-.9cm}
    \includegraphics[width=1.15\columnwidth]{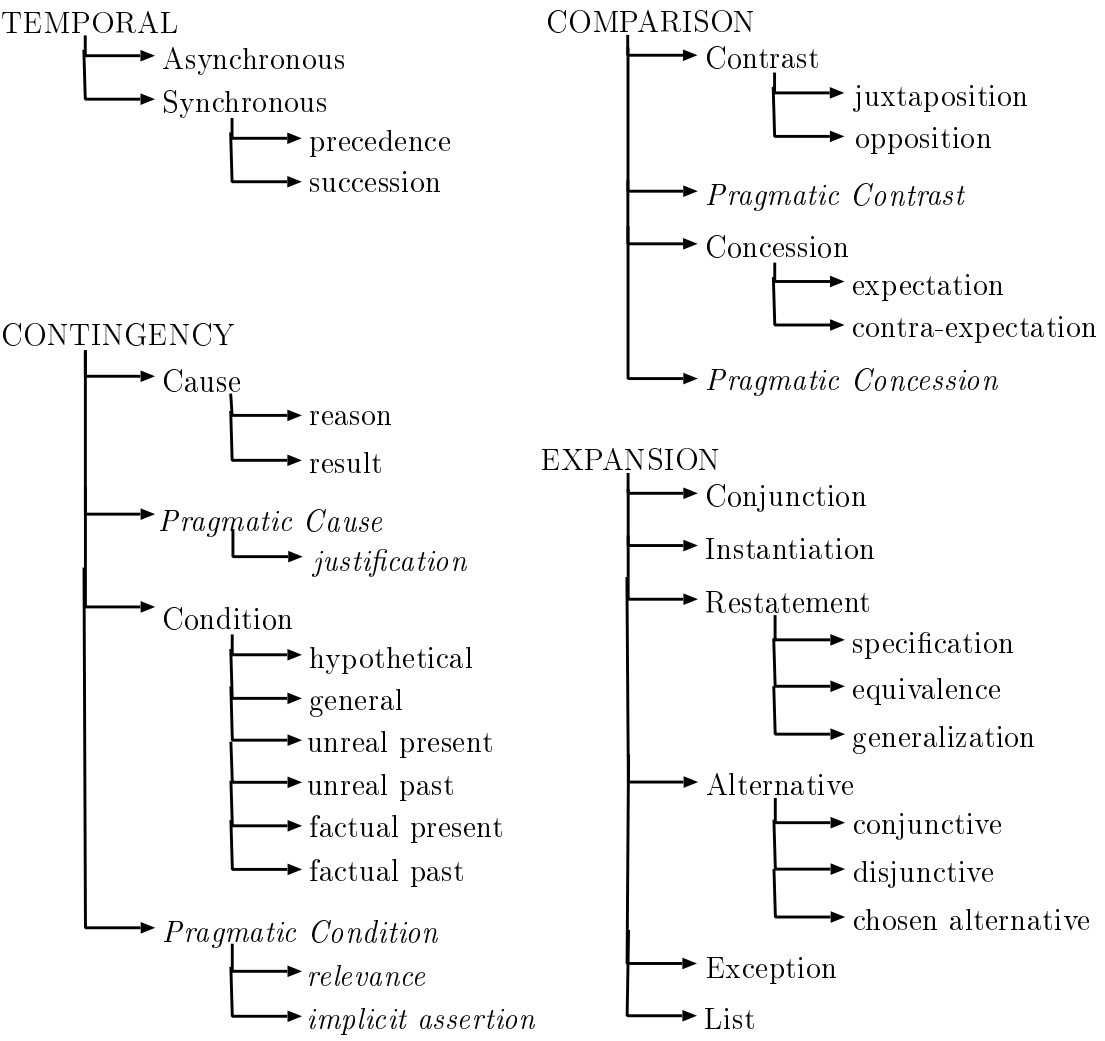}
    \caption{PDTB-2.0 Sense Tag Hierarchy.}
    \label{fig:q}
\end{figure}

\noindent The Penn Discourse TreeBank is a large corpus annotated for discourse relations and their arguments over the 1 million word Wall Street Journal corpus \cite{Prasad2008ThePD}. We use the second version of the PDTB in this paper in order to compare with prior literature. PDTB-2.0 expands the annotation  scheme, Figure \ref{fig:q}, found in PDTB-1.0 \cite{Prasad2006AnnotatingAI} by adding three levels: the highest level (\textit{Class}) containing four labels, the second (\textit{Type}) with 16, and the third with 23 possible \textit{Subtypes}. PDTB provides 15,604 implicit sentence pairs with one of these relation tags. In the current work, we focus on the highest level—that is, we predict \textit{Temporal}, \textit{Contingency}, \textit{Comparison}, and \textit{Expansion} relations. 

\subsection{Models}

Here, we outline the four models depicted in Figure \ref{fig:1} and how each incorporates additional context.

\begin{enumerate}
    \item Our Baseline model contains each implicit sentence pair and Level 1 (\textit{Class}) relation tag. 
    \item Our \textbf{D}irect \textbf{N}eighbors (DN) model contains every implicit sentence pair which has at least one implicit, explicit, or entity relation direct neighbor sentence; however, in training and testing, only implicit relation pairs were used. Here, we explored window size, testing both neighbors N=1 (9,427 observations) and N=2 (1,809 observations).
    \item Our \textbf{E}xpanded \textbf{W}indow \textbf{N}eighbors (EWN) model contains each implicit sentence pair with the addition of the nearest previous neighboring sentence from the same document and the nearest next sentence from the same document. Unlike DN which removes sentence pairs with no direct neighbors, due to unlabeled neighboring sentences, the EWN model combines the next nearest discourse unit, maintaining every sentence pair without filtration. Window sizes of N=2 and N=4 were tested for EWN.
    \item Our \textbf{P}art-\textbf{S}mart \textbf{R}andom \textbf{N}eighbors (PSRN) model contains each implicit sentence pair with the addition of a random prior neighboring sentence from the same document, and a random next sentence from the same document.
\end{enumerate}

Baseline, EWN, and PSRN all consisted of the same number of observations (15,604), DN was the only model with a reduced dataset as previously mentioned. The entire PDTB (i.e., implicit, explicit, and entity relations) was considered when concatenating additional context for EWN and PSRN, but again---as with DN---only pairs containing implicit relations were trained and tested on.

In creating these expanded context windows, we used the ‘SectionNumber’, ‘FileNumber’ and ‘SentenceNumber’ indexes to ensure that context windows were only created from the same document file.

\section{Methods}

\subsection{DistilBERT}

In this study, we use DistilBERT, a lightweight version of BERT \citep{Sanh2019DistilBERTAD} for discourse relation prediction. To measure the effects of expanded context windows on implicit sentence relation prediction, we utilize the \href{https://huggingface.co/}{Hugging Face} hyper-tuning pipeline to model our curated set-ups. Our initial model checkpoints came from the DistilBERT base uncased finetuned SST-2 text classification model available on the \href{https://huggingface.co/distilbert-base-uncased-finetuned-sst-2-english#model-details}{Hugging Face} website.

\subsection{Training Process}

\subsubsection{Text Encoding}  
\label{level3H}

Each sentence was tokenized and either padded or truncated to a length of 128 tokens. The sentences were then encoded with the \href{https://pypi.org/project/torch/}{Torch library} and converted into tensors along with their input IDs, attention masks, and labels to be our sentence embeddings.

\subsubsection{Fine-tuning on a Custom Dataset}  
\label{level3H}

The Hugging Face library allows for training on custom datasets provided they are in the right format structure. We created a `SententialNeighbors` class which inherits from torch.utils.data.Dataset and takes encoded input from the previous step and formats our data to fine-tune the DistilBERT model.


The dataset was split as advised by \citet{Prasad2006ThePD}, where Sections 0 through 22 were used for training, and Section 23 was reserved for testing. These parameters result in an approximate 95\% to training and 5\% to test split. 


\subsubsection{Model Setup and Training}  
\label{level3H}

We began our model design with the Hugging Face auto model for sequence classification. This is a generic model class instantiated with DistilBERT checkpoints for a sequence classification task with four output labels. To match our output dimension to the number of labels in our classification task, we added a final linear layer with an output dimension of four. Lastly, we configured our evaluation strategy to report on each epoch and set a random seed to 42 for reproducibility. 
In preliminary experiments, the addition of segment tags and retention of stop words was found to benefit the Baseline model. Thus, all models presented were run retaining both tags and stop words; each model was trained for 10 epochs.

\subsection{Evaluation}

\subsubsection{Statistical Tests}  
\label{level3H}

When comparing groups of evaluation results we ran independent two-sample t-tests. Because our samples were independent of each other and normally distributed, we could rely on these tests to determine statistical significance with an alpha p-value threshold set to 0.05. While accuracy, precision, recall, F1, and macro F1 were collected for each model run, we focused on accuracy and F1 as our primary evaluation metrics. 

\section{Results}

Overall, the Baseline model outperforms all proposed context window strategies. 

\begin{figure}[h]
    \centering
    \includegraphics[width=1.0\columnwidth]{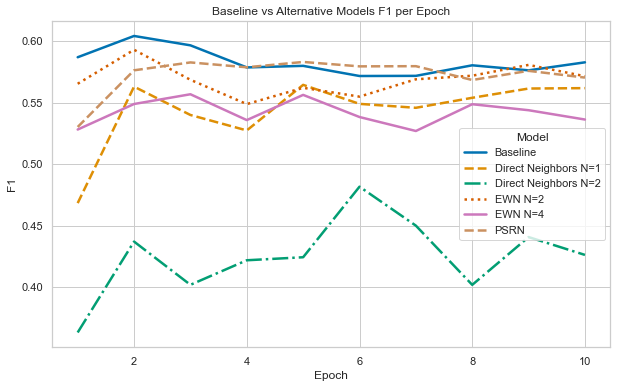}
    \caption{F1 Metrics for Baseline and All Models.}
    \label{fig_B-all}
\end{figure}

Figure 3 shows that the PSRN and EWN (N=2) models perform most competitively with Baseline. PSRN scarcely outperforms EWN (N=2) in accuracy (cf., PSRN at 58\% and EWN at 57.5\%). The difference in performance between the two strategies with regard to F1 is also almost negligible, with EWN (N=2) at 57.2\% and PSRN at 57\% by epoch 10. Further comparisons for precision, recall, and marco F1 may be observed in Table \ref{tab_B-all}.

\begin{table}[h]
\centering
\resizebox{1.0\columnwidth}{!}{%
\begin{tabular}{l|l|l|l|l|l}
\multicolumn{1}{r|}{\textit{Model}} & \multicolumn{1}{r|}{Accuracy}      & \multicolumn{1}{r|}{Precision}     & \multicolumn{1}{r|}{Recall}        & \multicolumn{1}{r|}{Macro F1}      & \multicolumn{1}{r}{F1}            \\ \hline
Baseline                            & \multicolumn{1}{r|}{\textbf{58.9}} & \multicolumn{1}{r|}{\textbf{58.3}} & \multicolumn{1}{r|}{\textbf{58.9}} & \multicolumn{1}{r|}{\textbf{47.9}} & \multicolumn{1}{r}{\textbf{58.3}} \\ \cline{1-1}
PSRN                                & \multicolumn{1}{r|}{58}            & \multicolumn{1}{r|}{56.6}          & \multicolumn{1}{r|}{58}            & \multicolumn{1}{r|}{42.9}          & \multicolumn{1}{r}{57}            \\ \cline{1-1}
EWN (N=2)                           & \multicolumn{1}{r|}{57.5}          & \multicolumn{1}{r|}{57}            & \multicolumn{1}{r|}{57.5}          & \multicolumn{1}{r|}{44.6}          & \multicolumn{1}{r}{57.2}          \\ \cline{1-1}
EWN (N=4)                           & \multicolumn{1}{r|}{54.2}          & \multicolumn{1}{r|}{53.5}          & \multicolumn{1}{r|}{54.2}          & \multicolumn{1}{r|}{40.5}          & \multicolumn{1}{r}{53.6}          \\ \cline{1-1}
DN (N=1)                            & \multicolumn{1}{r|}{56.8}          & \multicolumn{1}{r|}{56}            & \multicolumn{1}{r|}{56.8}          & \multicolumn{1}{r|}{44.8}          & \multicolumn{1}{r}{56.2}          \\ \cline{1-1}
DN (N=2)                            & \multicolumn{1}{r|}{48.7}          & \multicolumn{1}{r|}{38.1}          & \multicolumn{1}{r|}{48.7}          & \multicolumn{1}{r|}{32.5}          & \multicolumn{1}{r}{42.6}          \\ \hline  
\end{tabular}}
\caption{Metrics for Baseline and All Models on the 10th Epoch of training.}
\label{tab_B-all}
\end{table}

Table \ref{tab_related_work}, next page, situates the current work in relation to previous work. Bold indicates highest-performing overall, underlining indicates highest-performing in the present study. Notice that EWN (N=2) and DN (N=4) results are excluded, this is due to their overall poor performance (cf., Table \ref{tab_B-all}).

\begin{table*}[t] 
\centering
\resizebox{0.85\textwidth}{!}{%
\begin{tabular}{c|c|rr}
\multirow{2}{*}{\textbf{Model}} & \multirow{2}{*}{\textbf{Embeddings}} & \multicolumn{2}{c}{\textbf{Top-Level PDTB Relations}} \\ \cline{3-4} 
&& \multicolumn{1}{c|}{\textit{Macro avg F1}}& \multicolumn{1}{c}{\textit{Accuracy}} \\ \hline
LDSGM \cite{Wu2021ALD}& RoBERTa& \multicolumn{1}{r|}{\textbf{63.73}}& \textbf{71.18} \\ \cline{1-2}
BMGF-RoBERTa \cite{Liu2020OnTI}& RoBERTa& \multicolumn{1}{r|}{63.39}& 69.06 \\ 
\cline{1-2}
\citet{Bai2018DeepER}& ELMo& \multicolumn{1}{r|}{51.06}& - \\ 
\cline{1-2}
\citet{Dai2018ImprovingID}& word2vec& \multicolumn{1}{r|}{48.82}& 57.44 \\ 
\cline{1-2}
Baseline & DistilBERT & \multicolumn{1}{r|}{\underline{47.4}} & \underline{58.9} \\ 
\cline{1-2}
PSRN & DistilBERT & \multicolumn{1}{r|}{43.8} & 58 \\ 
\cline{1-2}
EWN (N=2) & DistilBERT & \multicolumn{1}{r|}{43.9} & 57.5 \\ 
\cline{1-2}
DN (N=1) & DistilBERT & \multicolumn{1}{r|}{40.2} & 56.8 \\ 
\hline
\end{tabular}}
\caption{Model Comparison to Related Work}
\label{tab_related_work}
\end{table*}

\subsection{Significance}

In this section, we observe significance between Baseline and context window strategies through accuracy and F1 measures (p < 0.05). These model comparisons show results from independent two-sample t-tests run per evaluation metric. In Table 3, the higher-performing model is marked in bold text.

\begin{table}[h]
\centering
\resizebox{1.05\columnwidth}{!}{%
\begin{tabular}{llllllll}
\multicolumn{1}{l|}{}& \multicolumn{3}{l|}{Accuracy}                              & \multicolumn{3}{l}{F1}                                                          &  \\ \cline{1-7}
\multicolumn{1}{l|}{\textit{\begin{tabular}[c]{@{}l@{}}Model 1\\ Model 2\end{tabular}}} & \multicolumn{1}{l|}{\begin{tabular}[c]{@{}l@{}}\textbf{Baseline}\\ PSRN\end{tabular}} & \multicolumn{1}{l|}{\begin{tabular}[c]{@{}l@{}}\textbf{Baseline}\\ EWN\end{tabular}} & \multicolumn{1}{l|}{\begin{tabular}[c]{@{}l@{}}\textbf{Baseline}\\ DN\end{tabular}} & \multicolumn{1}{l|}{\begin{tabular}[c]{@{}l@{}}\textbf{Baseline}\\ PSRN\end{tabular}} & \multicolumn{1}{l|}{\begin{tabular}[c]{@{}l@{}}\textbf{Baseline}\\ EWN\end{tabular}} & \begin{tabular}[c]{@{}l@{}}\textbf{Baseline}\\ DN\end{tabular} &  \\ \cline{1-7}
\multicolumn{1}{l|}{\textit{P-value}}                                                   & \multicolumn{1}{r|}{0.242}                                                   & \multicolumn{1}{r|}{0.014}                                                  & \multicolumn{1}{r|}{6.39E-05}                                              & \multicolumn{1}{r|}{0.096}                                                   & \multicolumn{1}{r|}{0.012}                                                  & \multicolumn{1}{r}{7.58E-04}                          &  \\ \cline{1-7}
\end{tabular}}
\caption{Significance across models.}
\label{tab_pval_B-all}
\end{table}

Baseline is also significantly different from both the EWN and DN context window strategies; however, Baseline is not significantly different from the PSRN strategy in accuracy (p = 0.242 > 0.05) nor in F1 (p = 0.096 > 0.05; see Table  3).

\section{Discussion}

In the context of the present work, the PSRN and EWN (N=2) results seem promising as these models are most competitive with the Baseline model; however, in the broad comparison to SOTA results, it appears that more context is detrimental to the task of implicit discourse relation classification.

The argument for why additional context should benefit the task of implicit discourse relation prediction is a linguistic one. As \citet{Dai2018ImprovingID} point out, the task can suffer from too narrow a scope. \citet{Atwell2021WhereAW}'s Wall Street Journal example, repeated below for convenience, demonstrates this point:

\begin{enumerate}
    \item \textit{\textbf{[}One housewife says: "With an electric kitchen I have to do my whole day's cooking the day before—and that during a couple of hours, not knowing from one minute to the next what time the power is coming on."\textbf{]}$_{A}$ \textbf{[}In this northern latitude it doesn't get dark in summer until about 10:30 p.m. so lighting is operate except at some crazy time like 11:45 at night, whenever there is power, unless they have stand-by diesel generators.\textbf{]}$_{B}$ \textbf{[}There's a year's supply of diesel oil here.\textbf{]}$_{C}$}
\end{enumerate}

The relation between B and C, without the context of A, may appear to be a \textit{Contingency} relation. The relation shift occurs because A adds the context of electricity issues to B. This contrasts with the initial reading of just B and C, where the relationship is that of B is as an unrealized situation which is \textit{contingent on} the situation in C. With the added context of A, the relation between B and C becomes a \textit{Comparison} (i.e., lighting is unreliable past 11:45 despite a year's supply of oil). 

Our negative results suggest different architectures may be better suited to the task. Approaches may also vary in how to approach the PDTB. The authors observed irregularities in PDTB's data structure, where some discourse units were doubly annotated, creating an one-to-many and many-to-one issue when concatenating unit pairs. PDTB is a hierarchy-based dataset and not necessarily temporal in every sense leading to the question whether PDTB was the best corpus to use for these experiments. For example, future work may attempt examining context in the RST Discourse Treebank \citep{Carlson2001BuildingAD}. 


Lastly, the size of our DN dataset somewhat inhibited our full exploration of the subject. Due to the confines of its very definition, direct neighbors requires adjacent discourse units. While the PDTB corpus is quite thorough in its annotations, we must accept the fact that not every pair of discourse units will necessarily have a discourse relation between them. Since some adjacent context is omitted, the DN dataset is smaller, which may have led to its poor performance. 

\section{Conclusion}

This paper set out to explore how context impacts the task of implicit discourse relation classification. We presented three models (DN, EWN, and PSRN) that slowly expanded the context involved in classification. Our experiments indicate that additional context is harmful in the task of implicit relation classification. Still, the authors hope that this work may inspire the curation and consideration of future datasets and how much context to include during automatic tagging.

\section{Ethics Statement} 

In this paper, the authors present three models aimed at exploring the effect of context in the task of implicit discourse relation classification. Since this work relies on annotated data, it is important to be mindful of where annotations come from in order to ensure both quality and integrity in how the they were curated. In these experiments, we used the Penn Discourse TreeBank, which may be accessed through the \href{https://catalog.ldc.upenn.edu/LDC2008T05}{Linguistic Data Consortium}. The authors do not observe any particular ethical concerns involved in the methods presented; however, we acknowledge that the task of implicit discourse relation classification may be an initial step for larger projects such as machine translation, natural language generation, or information extraction. 

\nocite{*}
\section{Bibliographical References}\label{sec:reference}
\bibliographystyle{lrec-coling2024_templates/lrec-coling2024-natbib}
\bibliography{bib}

\end{document}